%% file: main.tex
\documentclass{article}

\usepackage{microtype}
\usepackage{graphicx}
\usepackage{subfigure}
\usepackage{booktabs} 

\usepackage{mathtools} 
\usepackage[english]{babel}
\usepackage{algorithm}
\usepackage{hhline}
\usepackage{multirow}
\usepackage{array}
\usepackage{caption}
\usepackage[colorlinks=true,linkcolor=red,citecolor=blue]{hyperref}

\usepackage{pgfplots}
\pgfplotsset{compat=newest}
\usepgfplotslibrary{groupplots}
\usepgfplotslibrary{dateplot}
\input
\usepackage{makecell}
\makeatletter
\def\BState{\State\hskip-\ALG@thistlm}
\makeatother
\usepackage{setspace}

\makeatletter
\def\Cline#1#2{\@Cline#1#2\@nil}
\def\@Cline#1-#2#3\@nil{%
  \omit
  \@multicnt#1%
  \advance\@multispan\m@ne
  \Ifnum\@multicnt=\@ne\@firstofone{&\omit}\fi
  \@multicnt#2%
  \advance\@multicnt-#1%
  \advance\@multispan\@ne
  \leaders\hrule\@height#3\hfill
  \cr}
\makeatother

\makeatletter
\def\BState{\State\hskip-\ALG@thistlm}
\makeatother
\usepackage{colortbl}

\newcommand{\fade}[1]{\textcolor{gray}{#1}}

\graphicspath{{figures/}}
\usepackage{tabularx,adjustbox}
\usepackage{mathtools,commath,amsmath,amssymb,amsfonts,bm}\interdisplaylinepenalty=2500
\usepackage{amsthm}
\usepackage{multirow}
\usepackage{xspace}
\usepackage{enumitem} 

\newcolumntype{C}[1]{>{\hsize=#1\hsize\centering\arraybackslash}X}

\newcommand{\ie}{\textit{i.e.},\xspace}

\newtheorem*{theorem*}{Theorem}

\theoremstyle{definition}

\newtheorem{definition}{Definition}

\usepackage[accepted]{icml2021}

\icmltitlerunning{Collaborative Fairness and Adversarial Robustness in Federated Learning}

\begin{document}

\twocolumn[
\icmltitle{A Reputation Mechanism Is All You Need: Collaborative Fairness and Adversarial Robustness in Federated Learning}

\icmlsetsymbol{equal}{*}

\begin{icmlauthorlist}
\icmlauthor{Xinyi Xu}{equal,nus}
\icmlauthor{Lingjuan Lyu}{equal,ant}
\end{icmlauthorlist}

\icmlaffiliation{nus}{Department of Computer Science, University of Singapore, Singapore, Singapore}
\icmlaffiliation{ant}{Ant financial}

\icmlcorrespondingauthor{Xinyi Xu}{xinyi.xu@u.nus.edu}
\icmlcorrespondingauthor{Lingjuan Lyu}{lingjuanlvsmile@gmail.com}

\icmlkeywords{Machine Learning, ICML}

\vskip 0.3in
]

\printAffiliationsAndNotice{\icmlEqualContribution} 

\begin{abstract}
\textit{Federated learning}~(FL) is an emerging practical framework for effective and scalable machine learning among multiple participants, such as end users, organizations and companies.
However, most existing FL or distributed learning frameworks have not well addressed two important issues together: collaborative fairness and adversarial robustness~(e.g. free-riders and malicious participants). 
In conventional FL, all participants receive the global model~(equal rewards), which might be unfair to the high-contributing participants.
Furthermore, due to the lack of a safeguard mechanism, free-riders or malicious adversaries could game the system to access the global model for free or to sabotage it. 
In this paper, we propose a novel \emph{Robust and Fair Federated Learning}~(RFFL) framework to achieve collaborative fairness and adversarial robustness simultaneously via a reputation mechanism. RFFL maintains a reputation for each participant by examining their contributions via their uploaded gradients~(using vector similarity) and thus identifies non-contributing or malicious participants to be removed.
Our approach differentiates itself by \emph{not} requiring any auxiliary/validation dataset. Extensive experiments on benchmark datasets show that RFFL can achieve high fairness and is very robust to different types of adversaries while achieving competitive predictive accuracy.
\end{abstract}

\section{Introduction}
\textit{Federated learning}~(FL)~\cite{mcmahan2017communication} provides a promising collaboration paradigm by enabling a multitude of participants to construct a joint model without exposing their private training data. Two emerging challenges in FL are collaborative fairness (participants with different contributions should be rewarded differently), and adversarial robustness (free-riders 
should not enjoy the global model for free, and malicious participants should not compromise system integrity)~\cite{lyu2020threats}.

Most existing FL paradigms~\cite{mcmahan2017communication,kairouz2019advances,yang2019federated,li2019federated} allow all participants to receive the same model in the end regardless of their contributions~(by their uploaded parameters/gradients). 
This may lead to an unfair outcome as the participants who contribute the most are rewarded equally with the ones who contribute nothing.
In practice, there may be a number of reasons for why the contributions differ,
one reason is the divergence in the quality of the local data of different participants~\cite{zhao2019privacy}.
~\citep{FL2019} presented a motivating example for collaborative fairness: larger banks~(for fear of not being compensated fairly) may refuse to collaborate with smaller banks who have smaller client base and thus less high-quality data.

In terms of adversarial robustness, the conventional FL framework~\cite{mcmahan2017communication} is potentially vulnerable to adversaries and free-riders as it does not offer any safeguard mechanisms. The follow-up works considered robustness from different lens~\cite{blanchard2017machine,fung2020sybils,bernstein2019signsgd,yin2018byzantine}, but none of them can provide comprehensive supports for all the three types of attacks (targeted poisoning, untargeted poisoning and free-riders) considered in this work.

In summary, our contributions include:
\begin{itemize}
\item We propose a \emph{Robust and Fair Federated Learning}~(RFFL) framework to simultaneously achieve collaborative fairness and adversarial robustness. 
\item RFFL utilizes a reputation system to iteratively calculate participants' contributions and reward participants accordingly with different models of performance commensurate with their contributions.
\item Extensive experiments on benchmark datasets demonstrate that our RFFL can achieve high fairness and is very robust against all the investigated attacks (eg. targeted poisoning, untargeted poisoning and free-riders) while maintaining competitive predictive accuracy.
\end{itemize}

\section{Related work}
Promoting collaborative fairness has attracted substantial attention in FL.
One research direction uses incentive schemes combined with game theory, based on the rationale that participants should receive rewards commensurate with their contributions to incentivize good behaviour~\cite{Yang-et-al:2017IEEE,Gollapudi-et-al:2017,richardson2019rewarding,Yu-et-al:2020AIES}. Note that in all these works, all participants receive the same final model.

Another research direction addresses the egalitarian fairness notion, \ie equalizing the performance of all participants~\citep{mohri2019agnostic}, and a more generalized $q$-Fair FL~($q$-FFL)~\citep{Li2020fair}. The $q$-Fair gives participants with higher local losses higher weights~(optimizing their local objectives more relative to others).

As opposed to the above mentioned works, the most recent works~\citep{lyu2020towards,lyu2020-CFFL} are better aligned with collaborative fairness in FL, where model is used as rewards for FL participants, so the participants receive models of different performances commensurate with their contributions. \citet{lyu2020towards} adopted a mutual evaluation of local credibility mechanism, where each participant privately rates the other participants iteratively. However, their framework is mainly designed for a decentralized block-chain system, which may not be directly applicable to FL settings when a central server is deployed.
To alleviate this obstacle, \citet{lyu2020-CFFL} proposed a FL framework to achieve collaborative fairness via an additional validation dataset used by the server to determine the contributions of the participants.

In terms of robustness in FL, \citet{blanchard2017machine} proposed the Multi-Krum method based on a Krum function which excludes a certain number of uploaded gradients furthest from the mean and demonstrates resilience against up to 33\% \textit{Gaussian Byzantine} participants and up to 45\% omniscient Byzantine participants. \citet{fung2020sybils} presented FoolsGold to defend against Sybils. \citet{bernstein2019signsgd} proposed a communication efficient approach called SignSGD, which is robust to arbitrary scaling. In this approach, participants only upload the element-wise signs of the gradients without the magnitudes. A similar method was proposed by \citet{yin2018byzantine}, based on the statistics of the gradients, specifically element-wise median, mean and trimmed mean.

\section{Proposed RFFL Framework}
\label{sec:RFFL}
Our \emph{Robust and Fair Federated Learning}~(RFFL) framework focuses on two important goals in FL: \emph{collaborative fairness} and \emph{adversarial robustness}. We address both goals simultaneously via a reputation mechanism.

\textbf{Problem setting and notation.} We adopt the standard optimization model for FL: $\min_{\boldsymbol{w}\in \mathcal{W}} F(\boldsymbol{w}) \coloneqq \sum_{i=1}^{N} p_i F_i(\boldsymbol{w})$ where $N$ is the number of participants, $p_i$ is the weight of $i$-th participant such that $p_i\geq0$ and $\sum_{i=1}^N p_i =1$. $F_i(\cdot)$ is the respective local objective.
In round $t$, $\Delta \boldsymbol{w}^{(t)}_i \coloneqq \nabla F_i(\boldsymbol{w}^{(t-1)})$ and $ \Delta\boldsymbol{w}^{(t)} = \sum_{i=1}^N p_i \Delta \boldsymbol{w}^{(t)}_i$. $D$ denotes the number of parameters in the model $\boldsymbol{w}$.
$\text{cos}( \boldsymbol{u}, \boldsymbol{v}) = \langle  \boldsymbol{u}, \boldsymbol{v}  \rangle / (|| \boldsymbol{u}|| \times || \boldsymbol{v} ||)$ is the cosine similarity between two flattened gradient vectors.

\subsection{Collaborative Fairness}
The original FL framework~\citep{mcmahan2017communication} can be viewed as adopting the \textit{egalitarian} approach by giving everyone the same reward. \citet{mohri2019agnostic,Li2020fair} enforce the egalitarian concept through the lens of minimax optimization and fair resource allocation. However, the egalitarian approach might not be always desirable, if the participants are self-interested and not altruistic. For example in medicine, clinical trials data are time-consuming and expensive to collect, so researchers with limited resource may collaborate to conduct more extensive studies. Similar use cases for collaborative learning between competitors are present in finance~\citep{Yu-et-al:2020AIES}. As such, the participants are self-interested and ultimately competing against each other, so it is not desirable to share the final reward with everyone as it will not be \textit{fair} to the participants who have expended the most resources to collect data of higher qualities. If everyone is rewarded equally regardless of their contributions, then the participants may not have the motivation to contribute, by collecting data and uploading high-quality gradients.

The key idea for our reward design is that participants who contribute more should be rewarded better~\citep{Song2019-profit-allocation-FL,Wang2020-SV-in-FL,Hwee2020-icml}.
In addition to this qualitative relation, we propose a quantitative way to measure how `fair' a set of rewards are with respect to the contributions of the participants, via the Pearson correlation coefficient, $\rho_{\text{p}}(\cdot;\cdot)$.
To see this, consider the following 3-participant example: suppose the contributions are $\boldsymbol{v}=[1, 2, 10]$, and two possible sets of rewards $\boldsymbol{\phi}=[2,3,4]$ and $\boldsymbol{\phi}'=[2,4,20]$. 
Both $\boldsymbol{\phi}, \boldsymbol{\phi}'$ correctly reflect the qualitative relations in the contributions but intuitively $\boldsymbol{\phi}'$ is `fairer' as it better reflects the quantitative relations. This is indeed captured by the Pearson correlation coefficient where $\rho_{\text{p}}(\boldsymbol{v},\boldsymbol{\phi}) =0.9122 $ and $\rho_{\text{p}}(\boldsymbol{v}, \boldsymbol{\phi}') =1.0 $.
Note if $\boldsymbol{v}$ contains identical values, $\rho_{\text{p}}(\boldsymbol{v},\boldsymbol{\phi})$~(all participants contribute identically) is undefined, and in this case we reward all participants equally.

\begin{definition}[Collaborative Fairness]~\cite{lyu2020how,lyu2020-CFFL,lyu2020towards}\label{definition:collaborative-fairness}
Denote participants' real-valued contributions as $\boldsymbol{v}$, and a set of rewards as $\boldsymbol{\phi}$, then the quantitative collaborative fairness is: $\rho_{\text{p}}(\boldsymbol{v},\boldsymbol{\phi})$ where $\rho_{\text{p}}(\cdot;\cdot)$ is the Pearson correlation coefficient.
\end{definition}

\noindent\textbf{Choice of Contributions and Rewards.} 
The contributions $\boldsymbol{v}$ and rewards $\boldsymbol{\phi}$ should be quantitative and suitable to the FL setting. 
Intuitively, a participant with a larger and better dataset should be able to make higher contributions. 
Therefore, we adopt a simple empirical approach: the contributions are estimated by the standalone performance, \ie the test accuracy of a model trained only on the participant's \emph{local} data. Similarly, the rewards are represented by the final performance, \ie the test accuracy of the model received by the participant at the end of FL. While in the original FL formulation~\citep{mcmahan2017communication}, the reward is the same for everyone because all participants synchronize with the server.
In order to achieve collaborative fairness, we require the models to have different performance, commensurate with their contributions.

\subsection{Adversarial Robustness}
For adversarial robustness, we consider the threat model in Definition~\ref{def:byz}~\citep{blanchard2017machine,yin2018byzantine}. 
\begin{definition}[Threat Model]
\label{def:byz}
In the $t$-th round, an honest participant uploads $\Delta\boldsymbol{w}_i^{(t)}\coloneqq\nabla F_i(\boldsymbol{w}_i^{(t)})$ while a dishonest participant/adversary can upload arbitrary values. 
\end{definition}

In particular, we investigate three types of attacks: 
\begin{itemize}[noitemsep,nolistsep]
    \item \textbf{\textit{Targeted poisoning}}. We consider the label-flipping attack, in which the labels of training examples are flipped to a target class~\cite{biggio2011support}. For instance, in MNIST a `1-7' flip refers to training on images of `1' but using `7' as the labels. 
    \item \textbf{\textit{Untargeted poisoning}}. We consider three types of untargeted poisoning defined in~\cite{bernstein2019signsgd}, where before uploading gradients, the adversary may (i) arbitrarily rescale gradients; or (ii) randomize the element-wise signs of the gradients; or (iii) randomly invert the element-wise values of the gradients.
    \item \textbf{\textit{Free-riders}}. Free-riders represent the participants unwilling to contribute their gradients due to data privacy concerns or computational costs, but want to access the jointly trained model for free~\cite{FL2019}. They typically upload random gradients. 
\end{itemize}

\subsection{Robust and Fair FL~(RFFL) via Reputation}

\textbf{Intuition for RFFL.}
The key to achieving collaborative fairness and adversarial robustness is in the uploaded gradients from the participants.
A high-quality gradient~(trained on high-quality local data) carries useful information for participants while a low-quality and possibly adversarial gradient can impair model performance.
In gradient-based learning, a high-quality gradient moves the model towards lower loss quickly, while a low-quality or adversarial gradient can move the model very slowly or even move the model towards higher loss.

In particular, for a participant $i$, we manage the gradients $i$ downloads so that $i$'s model can move towards lower loss~(mitigate adversarial gradients) at a rate commensurate with the quality of $i$'s uploaded gradients~(achieve collaborative fairness).
To do so, we propose an iteratively updated \textit{reputation} for each participant. This reputation is maintained by the server, and not seen by the participants.

\textbf{High-level overview.}
RFFL makes two important modifications to the conventional FL framework: in the gradient aggregation, and in the downloading of the gradients for the participants. In addition, by keeping a reputation for each participant and a pre-determined threshold, we can achieve collaborative fairness~(rewarding participants commensurately according to their reputations) and adversarial robustness~(identifying and removing adversaries).
The detailed realization of RFFL is given in Algorithm~\ref{Algorithm:Fair_FL}. Our code is available at: https://github.com/XinyiYS/Robust-and-Fair-Federated-Learning

\begin{algorithm}[ht]
\caption{Robust and Fair Federated Learning
(RFFL)}\label{Algorithm:Fair_FL}
\small
\begin{algorithmic}[1]
\STATE {\bfseries Input:} moving average coefficient $\alpha$, reputation threshold $\beta$; gradient normalizing constant $\gamma$. 
\STATE {\bfseries Notations:} 
${r}_i^{(t)}$ is $i$'s reputation in round $t$ ; $R \coloneqq \{i | {r}_i^{(t)} \geq \beta\}$ is the reputable set and w.l.o.g $\sum_{i \in R} {r}_i^{(t)} = 1$; $\boldsymbol{w}_i$ and $\boldsymbol{w}_g$ denote participant $i$ and server model parameters, respectively
     \STATE {\hskip 8em \bfseries Participant $i$}  
    \STATE Upload local gradients $\Delta \boldsymbol{w}_i^{(t)}\coloneqq \nabla F_i(\boldsymbol{w}_i^{(t)})$ to server
    \STATE Download the allocated gradients $\Delta \boldsymbol{w}_{*i}^{(t)}$, and integrate with local gradients:
    \STATE $\boldsymbol{w}_i^{(t+1)}= \boldsymbol{w}_i^{(t)}  + \Delta \boldsymbol{w}_i^{(t)} + \Delta \boldsymbol{w}_{*i}^{(t)}$ 
    \vspace{2mm}
    
    \STATE {\hskip 8em \bfseries Server}\\
    \emph{Aggregation}:
    \STATE $\Delta \boldsymbol{w}_g^{(t)} = {\textstyle\sum}_{i \in R}  {r}_i^{(t-1)} 
        \Delta\boldsymbol{w}_i^{(t)} \times \gamma /|| \Delta\boldsymbol{w}_i^{(t)} ||    $ 
    \FOR{$i \in R$}
     \STATE $\tilde{r}^{(t)}_i = \text{cos}(\Delta\boldsymbol{w}_g^{(t)}, \Delta \boldsymbol{w}_i^{(t)})$
     \STATE $r^{(t)}_i = \alpha r^{(t-1)}_i + (1-\alpha ) \tilde{r}^{(t)}_i$
    \IF{${r}_i^{(t)}<\beta$}
    \STATE $R = R\setminus \{i\}$ \fade{Remove too low reputations}
    \ENDIF
    \ENDFOR\\
    \emph{Download}:
    \FOR{$i \in R$} 
    \STATE $\text{quota}_i =  D \times {r}_i^{(t)} /(\max_i   r^{(t)}_i) $
    \STATE $\Delta \boldsymbol{w}_{*i}^{(t)} = \texttt{sparsify}(\Delta \boldsymbol{w}_g^{(t)} , \text{quota}_i) - {r}_i^{(t-1)} \Delta \boldsymbol{w}_i^{(t)}$ 
    \ENDFOR
\end{algorithmic}
\end{algorithm}

\textbf{Aggregation step.} During gradient aggregation step, the server adopts reputation-weighted aggregation: 
\begin{equation}\label{equ:aggregate}
    \Delta\boldsymbol{w}_g^{(t)}=  {\textstyle\sum}_{i \in R}{r}_i^{(t-1)}\Delta\boldsymbol{w}_i^{(t)} \times  \gamma / ||\Delta\boldsymbol{w}_i^{(t)}||
\end{equation}
where $\Delta\boldsymbol{w}_i^{(t)} \coloneqq \nabla F_i(\boldsymbol{w}_i^{(t-1)})$ is the uploaded gradient by participant $i$ and $\gamma$ is a normalization coefficient to prevent gradient explosion~\cite{Lin2018-deep-gradient-compression-momentum-correction,Pascanu13-normalization-of-gradient}. $R$ is the set of reputable participants, i.e., those whose reputations are higher than a pre-determined threshold $\beta$. The reputation for each round is calculated as follows,
\begin{equation}\label{equ:reputation}
r^{(t)}_i = \alpha r^{(t-1)}_i + (1-\alpha ) \tilde{r}^{(t)}_i  
\end{equation}
where $\tilde{r}^{(t)}_i = \text{cos}(\Delta\boldsymbol{w}_g^{(t)}, \Delta \boldsymbol{w}_i^{(t)})$ is $i$'s reputation in the current round and $\alpha$ is a settable weight coefficient.
Cosine similarity has previously been used in determining the quality of gradients to improve model performance~\cite{Cao2020-FLTrust,fung2020sybils}.
In~(\ref{equ:reputation}), we integrate the reputation in both the current round and the previous round, in order to update the reputations in a smooth way and mitigate noise incurred by the training process and random model initializaiton~\cite{Song2019-profit-allocation-FL, Wang2020-SV-in-FL}.

\textbf{Download step.} During the download step, the server determines the gradient $i$ can download based on $r^{(t)}_i$ as follows,
\begin{equation}\label{equ:download}
    \Delta \boldsymbol{w}_{*i}^{(t)} = \texttt{sparsify}(\Delta \boldsymbol{w}_g^{(t)}, \text{quota}_i) - r_i^{(t-1)} \Delta \boldsymbol{w}_i^{(t)}
\end{equation}
where $\text{quota}_i = D \times  r_i^{(t)}/ (\max_j r_j^{(t)})$ is the number of parameters to be downloaded and determined by the relative reputation, $r_i^{(t)} / (\max_j r_j^{(t)})$.
After $\text{quota}_i$ is calculated, the server first constructs a sparsified version of the aggregated gradient $\Delta \boldsymbol{w}_g^{(t)}$ by retaining only the largest $\text{quota}_i$ values, then
removes $i$'s own gradient from it.
$\Delta \boldsymbol{w}_{*i}^{(t)}$ refers to the gradient for $i$ to download.

Sparsifying gradient vectors by retaining only the largest values gradually reduces the information and thus the quality of the gradient~\citep{Alistarh2018-convergence-sparsified-gradients,Yan2020-dual-way-gradient}, which allows us to design rewards based on the contributions of the participants.
Simply put, $i$ with a higher contribution downloads a less sparsified gradient.

\textbf{Reputation threshold} $\beta$\textbf{.}
We introduce a reputation threshold $\beta$ as a settable coefficient to impose a requirement for the least amount of contribution from the participants.
It can also be used to identify and remove adversaries as their contributions are usually low.
In each round $t$, the updated reputations $r_i^{(t)}$ are compared against $\beta$ and participants with reputations less than $\beta$ are removed from the subsequent rounds. 
Specifically, $R$ denotes the participants with reputations higher than $\beta$. The removed participants will \textit{not} be added back in later.


\section{Experiments}
\label{sec:Performance}
\subsection{Datasets}
We conduct experiments on various datasets including: (1) image classifications datasets: MNIST~\citep{lecun1998gradient} and CIFAR-10~\citep{krizhevsky2009learning}; (2) text classifications datasets: Movie review~(MR)~\citep{pang2005seeing} and Stanford sentiment treebank~(SST)~\citep{kim2014convolutional}. We use a 2-layer convolutional neural network~(CNN) for MNIST~\cite{cnn-mnist}, a 3-layer CNN for CIFAR-10~\citep{cnn-imagenet} and a text embedding CNN for MR and SST~\citep{kim2014convolutional}.

\subsection{Baselines}
We examine performance via three metrics : 1) predictive performance; 2) fairness and 3) robustness. 
For predictive performance, we include FedAvg~\cite{mcmahan2017communication}, and the \textit{Standalone} framework where participants train locally without collaboration.
For fairness performance, we focus our comparison with $q$-FFL~\cite{Li2020fair} and CFFL~\cite{lyu2020-CFFL}.
For robustness performance, we include several Byzantine-tolerant and/or robust FL frameworks including Multi-Krum~\cite{blanchard2017machine}, FoolsGold~\cite{fung2020sybils}, SignSGD~\cite{bernstein2019signsgd} and Median~\cite{yin2018byzantine}. FoolsGold was designed to mitigate sybils attacks and we adapt it for comparison.

\subsection{Experimental Setup}
\label{sec:Setup}

\textbf{Data splits.}
In addition to the standard I.I.D data sampling regime~(`uniform' split, denoted as UNI), we consider two heterogeneous data splits by varying the data set sizes and the class numbers respectively.
We follow a power law to randomly partition total \{3000,6000,12000\} MNIST examples among \{5,10,20\} participants respectively. In this way, each participant has a distinctly different number of examples, with the first participant has the least and the last participant has the most~(on average 600~\cite{mcmahan2017communication}). We refer to this as the `powerlaw' split~(POW). Data splits for CIFAR-10, MR and SST datasets follow a similar way, with details in the appendix.
Next, we investigate `classimbalance' split~(CLA), for which we vary the number of distinct classes in each participant's dataset, increasing from the first participant to the last. For example, for MNIST with total 10 classes and 5 participants, participant-\{1,2,3,4,5\} owns \{1,3,5,7,10\} classes of digits respectively.
All participants have the same data size, but different class numbers. We only investigate MNIST and CIFAR-10 dataset as they both contain $10$ classes.

\noindent\textbf{Adversaries.} We consider three types of adversaries on MNIST: targeted poisoning as in label-flipping~\cite{biggio2011support}, untargeted poisoning as in the blind multiplicative adversaries~\cite{bernstein2019signsgd}, and free-riders. In each experiment, we evaluate RFFL against one type of adversary, and we test two proportions of the adversaries~(to the honest participants), 20\% and 110\%. For targeted poisoning, the adversary uses `7' as labels for actual `1' images, during their local training to produce `crooked' gradients. For untargeted poisoning, we consider three sub-cases separately, the adversary: 1) re-scales the gradients by $-100$; 2) randomizes the element-wise signs; or 3) randomly takes the element-wise reciprocals. For free-riders, they upload gradients randomly drawn from the $[-1, 1]$ uniform distribution. We conduct experiments with adversaries under UNI and POW for illustration purpose.

\noindent\textbf{Hyper-Parameters.} We set the reputation threshold to $\beta= 1 / (3N)$, the moving average coefficient $\alpha = 0.95$ and the gradient normalizing constant $\gamma=0.5$ for MNIST, $0.15$ for CIFAR-10, and $1$ for MR and SST. The interpretation for $\beta = 1/ (3N)$ is that each participant~(in order not to be removed from $R$) should contribute at least $1/3$ of their individual proportion which is $1/N$ as there are $N$ participants.
Further details on hyperparameters, hardware resource, and runtime statistics are included in Appendix~\ref{sec:A1}. 

\subsection{Experimental Results}
\label{sec:results}

\noindent\textbf{Predictive performance.}
Table~\ref{tbl:accuracy} reports the average and maximum accuracy of participants' final local models. 
RFFL outperforms other methods by a noticeable margin, especially for more heterogeneous data splits.
It may be attributed to the reputation-weighted aggregation which can dynamically up-weight the participants who contribute more~(implying they have better local data)~\cite{Li2020On-fedavg-noniid}.

\noindent\textbf{Collaborative Fairness.}
Table~\ref{tbl:fairness} shows the calculated fairness results~(the Pearson correlation coefficient between the standalone performance and the final performance).
We use the standalone performance because it can estimate the contributions of the participants and more importantly because it is independent of the methods so can be used to compare `fairness' results across methods.  
The results indicate that in RFFL, participants who have better local data~(contribute more) get better models. Fig.~\ref{fig:mnist_cifar10_all} in Appendix~\ref{sec:A2} provides an illustration where in MNIST and CIFAR-10, the agent with larger index has better final performance~(because their local data are better in quantity and/or quality, under POW and CLA).
While CFFL outperforms RFFL in some cases, CFFL requires an additional auxiliary dataset for validation, to determine the contributions of the participants.
The results for the 5-participant case on both MNIST and CIFAR-10 are included in Appendix~\ref{sec:A2}.

\begin{table*}[ht]
\caption{Average and Maximum Test Accuracy[\%]. Values in brackets denote maximum accuracy among $N$ participants.
}
\small
\label{tbl:accuracy}
\centering
\begin{tabularx}{\linewidth}{|l|*{3}l|*{3}l|*{3}l|*{1}l|*{1}{C{1}}|}
\hline
\multirow{1}{*}{} & \multicolumn{6}{c|}{MNIST} & \multicolumn{3}{c|}{CIFAR-10} & \multicolumn{1}{c|}{MR} & 
\multicolumn{1}{c|}{SST}
\tabularnewline
\hline
\multirow{1}{*}{$N$} & \multicolumn{3}{c|}{10} & \multicolumn{3}{c|}{20} & \multicolumn{3}{c|}{10} & \multicolumn{1}{c|}{5} & 
\multicolumn{1}{c|}{5}
\tabularnewline
\hline
Data Split
 & UNI & POW & CLA & UNI & POW & CLA
 & UNI & POW & CLA & POW & POW
\tabularnewline
\hline
\textit{Standalone} 
&91(91) & 88(92) & 53(92) &91(91) &  89(92) &  48(90) 
&46 (47) & 43 (49) & 31 (44)
&47(56) &31(34)
\tabularnewline
\hline
FedAvg
& 93(94) & 92(94) & 53(93) & 93(93) &  92(94) &  49(92)
& 48 (48) & 47 (50) & 32 (47) 
&51(63)&33(35)
\tabularnewline
$q$-FFL
&  85(91) &  27(45) &  44(64) &  88(91) &  48(53) &  40(59)
&  41 (46) &  36 (36) &  22 (28)
&  12(18) 
&  23(25)
\tabularnewline
CFFL
&  90(92) &  85(90) &  34(44) 
&  91(93) &  88(91) &  39(46)
&  39 (41) &  35 (45) &  22 (40)
&  44(53)  
& 31(32)
\tabularnewline
\hline
RFFL
& \textbf{96(96)} & \textbf{95}(\textbf{96}) & \textbf{73(94)} & \textbf{97}(\textbf{97}) &  \textbf{95(96)} &  \textbf{66}(\textbf{95})
&  \textbf{61 (62)} &  \textbf{59 (61)} &  \textbf{35 (54)}
& \textbf{57(76)}
& \textbf{35(37)}
\tabularnewline
\hline
\end{tabularx}

\caption{Fairness results[\%] as in Definition~\ref{definition:collaborative-fairness} between the standalone performance and the final performance of $N$ participants. 
}
\small
\label{tbl:fairness}
\centering
\begin{tabularx}{\linewidth}{|l|*{3}l|*{3}l|*{3}l|*{1}l|*{1}{C{1}}|}
\hline
\multirow{1}{*}{} & \multicolumn{6}{c|}{MNIST} & \multicolumn{3}{c|}{CIFAR-10} & \multicolumn{1}{c|}{MR} & 
\multicolumn{1}{c|}{SST}
\tabularnewline
\hline
\multirow{1}{*}{$N$} & \multicolumn{3}{c|}{10} & \multicolumn{3}{c|}{20} & \multicolumn{3}{c|}{10} & \multicolumn{1}{c|}{5} & \multicolumn{1}{c|}{5}
\tabularnewline
\hline
Data Split
 & UNI & POW & CLA & UNI & POW & CLA
 & UNI & POW & CLA & POW & POW
\tabularnewline
\hline
FedAvg
& $-$31.2  &  77.33  & 64.53
& 3.85    &  $-$3.58 &  70.83
& $-$42.9 & 40.58 & 79.34
& 22.22 &  64.18
\tabularnewline
$q$-FFL
&  -44.73 &    39.00 &    22.38 &   -22.01 &    38.71 &    48.07
&  -17.64 &    51.33 &    94.06
&  56.43 &  -75.92
\tabularnewline
CFFL
&   \textbf{83.57} &    91.80 &    81.24 &  \textbf{ 82.52} &    94.70 &    85.71
&   78.25 &    72.55 &    81.31
&   96.85 &   \textbf{93.34}
\tabularnewline
\hline
RFFL
& 83.36 & \textbf{98.33}  & \textbf{99.81} & 75.19  & \textbf{97.88} &  \textbf{99.64}
& \textbf{81.93} & \textbf{98.78}  & \textbf{99.89}
& \textbf{99.59} & 65.88
\tabularnewline
\hline
\end{tabularx}
\end{table*}

\noindent\textbf{Adversarial robustness.}
We first demonstrate RFFL's effectiveness in identifying and isolating the untargeted poisoning adversaries and free-riders as shown in Figs.~\ref{fig:mnist_fr_reputations} and~\ref{fig:mnist_ma_reputations} in Appendix~\ref{sec:A2}.
The figures show the reputations of free-riders and the untargeted poisoning adversaries quickly decrease to below the threshold $\beta = 1 /(3N)$ and get removed from subsequent rounds.
For targeted poisoning, the adversaries are not completely identified. It is possible since an adversary intentionally mislabelling only one digit out of ten may still meet the reputation threshold of $\beta = 1/(3N)$.

For targeted poisoning, we consider two additional metrics~\cite{fung2020sybils}: targeted class accuracy and attack success rate. Targeted class accuracy in our experiment corresponds to the test accuracy on digit `1' images. Attack success rate corresponds to the proportion of `1' images incorrectly classified as `7'.
The results are in Tables~\ref{tbl:targeted_poisoning} and~\ref{tbl:targeted_poisoning_11}.
Table~\ref{tbl:targeted_poisoning} illustrates that FedAvg, Multi-Krum and RFFL perform well in all three metrics. FedAvg and Multi-Krum are robust against 20\% label flipping adversaries because these introduced `crooked' gradients that are outweighed by the gradients from the honest participants. RFFL performs well by reducing the negative effect from these adversaries. 

Somewhat surprisingly, the robustness was minimally affected when there are 10 honest participants and 11 adversaries~( Table~\ref{tbl:targeted_poisoning_11}). Additional results in Appendix~\ref{sec:A2} indicate that, with an overwhelming number of adversaries, the honest participants can be mistaken as `adversaries', so the honest participants receive gradually lower reputations and eventually get removed. Afterwards, the honest participants conduct training alone and will not be affected by adversaries. Therefore, under this attack, our method does have a limit to the number of adversaries beyond. This in a way agrees with common methods which assume the majority of all participants are honest~\citep{blanchard2017machine}. The challenge is how to reliably detect these label-flipping adversaries who provide meaningful gradients~(since they are only intentionally mis-classifying one class out of ten), from honest participants who may simply have smaller or lower quality data. We leave a more in-depth investigation to future work.

For untargeted poisoning, the results are in Tables~\ref{tbl:untargeted_sign_uniform},~\ref{tbl:untargeted_rescaling_uniform} and~\ref{tbl:untargeted_inverting_uniform}. 
These results demonstrate that RFFL is overall the most robust.
We observe that Multi-Krum and FoolsGold are not robust against untargeted poisoning. Multi-Krum utilizes the mean vector of the gradients, and is thus not robust to rescaling attacks. FoolsGold was designed to be robust against adversaries with a common objective, which is \textit{not} the case for untargeted poisoning. Both SignSGD and Median demonstrate some degree of robustness for re-scaling attack. SignSGD is robust against re-scaling attack as it preserves the signs of gradients. Median utilizes the median statistic and is robust against extreme outliers as in re-scaling attack.

For the free-rider scenario, only FedAvg and RFFL are consistently robust. FedAvg is robust because the gradients from the free-riders have an expected value of zero, so the additional noise does not affect the asymptotic unbiasedness. Among the others, Multi-Krum exhibits some degree of robustness but compromises the accuracy. FoolsGold is not robust against free-riders as it assumes that the honest participants produce gradients that are more random than the adversaries who share a common attack objective function. For SignSGD, the free-riders are exactly the sign-randomizing adversaries, so the behavior is consistent. For Median, it is possible that the honest gradients are small and thus close to random noisy gradients, and as a result the random noisy gradients get updated to the model.

Additional experiments for robustness under POW are in Appendix~\ref{sec:A2}.

\begin{table}[ht]
\caption{Maximum accuracy [\%], Attack success rate [\%] and Target accuracy [\%] for MNIST under UNI with 10 honest participants and additional 20\% \textbf{label-flipping} adversaries.}
\label{tbl:targeted_poisoning}
\centering
\small
\begin{tabularx}{\linewidth}{|l|*{3}{C{1}|}}
\hline
{} &  Max accuracy & Attack success rate & Target accuracy \\
\hline
FedAvg  	&  \textbf{96.8}  &  0.2 &  98.8 \\
\hline
FoolsGold 	&  9.8 &  \textbf{0} &  0 \\
\hline
Multi-Krum  &  95.6 &  0.2 &  \textbf{99.0}  \\
\hline	
SignSGD  	&  9.1 &  41.9 &  18.8  \\
\hline
Median    	&  0.3 &  0.5 &  0.1  \\
\hline
RFFL  		&  93.4 &  \textbf{0} & 98.9  \\
\hline
\end{tabularx}
\end{table}

\begin{table}[!ht]
\caption{Maximum accuracy [\%], Attack success rate [\%] and Target accuracy [\%] for MNIST under UNI with 10 honest participants and additional 110\% \textbf{label-flipping} adversaries. 10 honest participants and 11 adversaries.}
\label{tbl:targeted_poisoning_11}
\centering
\small
\begin{tabularx}{\linewidth}{|l|*{3}{C{1}|}}
\hline
{} &  Max accuracy & Attack success rate & Target accuracy \\
\hline
FedAvg  	&  90.9 &  48.6 &  49.3 \\
\hline
FoolsGold 	&  19.2 &  \textbf{0} &  55.0\\
\hline
Multi-Krum  &  \textbf{96.3} &  0 &  98.8  \\
\hline	
SignSGD  	&  9.1 &  0 &  18.8  \\
\hline
Median    	& 8.2 &  0 &  72.3  \\
\hline
RFFL  		&  93.5 &  \textbf{0} & \textbf{99.1}  \\
\hline
\end{tabularx}
\end{table}

\begin{table}[!ht]
\caption{Individual test accuracies [\%] over MNIST under UNI with 10 honest participants and additional 20\% \textbf{sign-randomizing} adversaries. Adversaries omitted.}
\centering
\small
\label{tbl:untargeted_sign_uniform}
\begin{tabularx}{\linewidth}{|l|*{11}{C{1}|}}
\hline
{} &  1 &  2 &  3 &  4 &  5 &  6 &  7 &  8 &  9 & 10 \\
\hline
FedAvg     & 10 & 10 & 10 & 10 & 10 & 10 & 10 & 10 & 10 &  10 \\
\hline
FoolsGold  & 11 & 11 & 11 & 11 & 11 & 11 & 11 & 11 & 11 &  11 \\
\hline
Multi-Krum & 10 & 10 & 10 & 10 & 10 & 10 & 10 & 10 & 10 &  10 \\
\hline
SignSGD    &  9 &  9 &  9 &  9 &  9 &  9 &  9 &  9 &  9 &   9 \\
\hline
Median     &  1 &  1 &  1 &  1 &  1 &  1 &  1 &  1 &  1 &   1 \\
\hline
RFFL       & 92 & 92 & 94 & 91 & 92 & 93 & 92 & 92 & 92 &  92 \\
\hline
\end{tabularx}
\end{table}

\begin{table}[!ht]
\caption{Individual test accuracies [\%] over MNIST under UNI with 10 honest participants and additional 20\% \textbf{re-scaling} adversaries. Adversaries omitted.}
\centering
\small
\label{tbl:untargeted_rescaling_uniform}
\begin{tabularx}{\linewidth}{|l|*{11}{C{1}|}}
\hline
{} &  1 &  2 &  3 &  4 &  5 &  6 &  7 &  8 &  9 & 10 \\
\hline
FedAvg     &  10 &  10 &  10 &  10 &  10 &  10 &  10 &  10 &  10 &   10 \\
\hline
FoolsGold  &  10 &  10 &  10 &  10 &  10 &  10 &  10 &  10 &  10 &   10 \\
\hline
Multi-Krum &  10 &  10 &  10 &  10 &  10 &  10 &  10 &  10 &  10 &   10 \\
\hline
SignSGD    &  50 &  58 &  62 &  58 &  59 &  64 &  66 &  57 &  57 &   57 \\
\hline
Median     &  11 &  10 &  39 &  28 &  20 &  40 &  48 &  27 &  35 &   28 \\
\hline
RFFL       &  93 &  93 &  94 &  92 &  92 &  94 &  94 &  93 &  93 &   92 \\
\hline

\end{tabularx}
\end{table}

\begin{table}[ht]
\caption{Individual test accuracies [\%] over MNIST under UNI with 10 honest participants and additional 20\% \textbf{value-inverting} adversaries. Adversaries omitted.}
\centering
\small
\label{tbl:untargeted_inverting_uniform}
\begin{tabularx}{\linewidth}{|l|*{11}{C{1}|}}
\hline
{} &  1 &  2 &  3 &  4 &  5 &  6 &  7 &  8 &  9 & 10 \\
\hline
FedAvg     &  9 &  9 &  9 &  9 &  9 &  9 &  9 &  9 &  9 &   9 \\
\hline
FoolsGold  &  8 &  8 &  8 &  8 &  8 &  8 &  8 &  8 &  8 &   8 \\
\hline
Multi-Krum & 17 & 17 & 17 & 17 & 17 & 17 & 17 & 17 & 17 &  17 \\
\hline
SignSGD &  9 &  9 &  9 &  9 &  9 &  9 &  9 &  9 &  9 &   9 \\
\hline
Median     &  1 &  1 &  1 &  1 &  1 &  1 &  1 &  1 &  1 &   1 \\
\hline
RFFL       & 92 & 93 & 94 & 92 & 92 & 93 & 93 & 93 & 93 &  92 \\
\hline
\end{tabularx}
\end{table}

\section{Discussion \& Conclusion}
We propose a Robust and Fair Federated Learning (RFFL) framework to address both \textit{collaborative fairness} and \textit{adversarial robustness} in FL. 
RFFL achieves these two goals by introducing reputations and iteratively evaluating the contribution of each participant, via the cosine similarity between the uploaded local gradients and the aggregated global gradients. 
Extensive experiments on various datasets demonstrate that RFFL achieves higher accuracy than FedAvg, and is robust against various types of adversaries under various settings. 
For future work, we plan to explore and theoretically formalize the potential trade-off among these three metrics: predictive performance, collaborative fairness and adversarial robustness.

\newpage
\bibliography{biblio}
\bibliographystyle{icml2021}

\newpage
\appendix
\input{appendix}

\end{document}

%% file: appendix.tex
\section{Additional Experimental Results}

\begin{figure*}[!t]
\centering
        \begin{subfigure}{\em}
                \includegraphics[width=4.5cm,height=4.2cm]{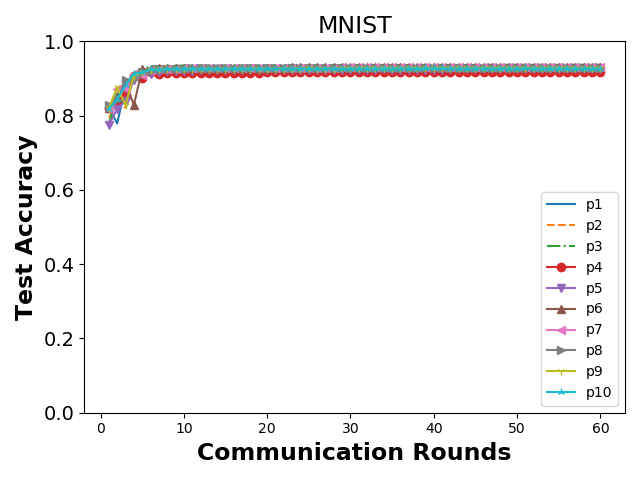}
        \end{subfigure}
        \begin{subfigure}{\em}
                \includegraphics[width=4.5cm,height=4.2cm]{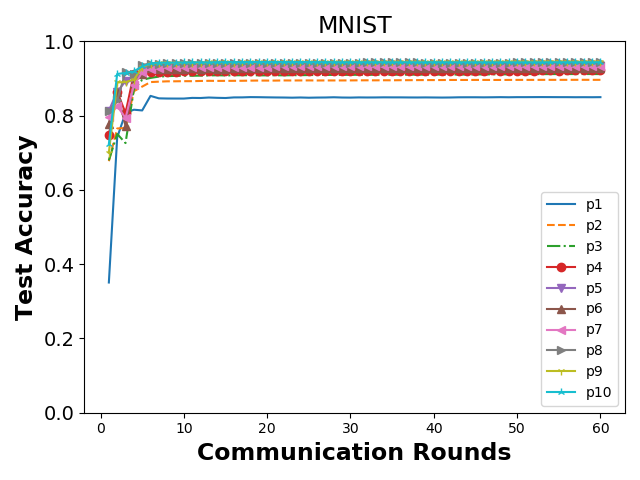}        
        \end{subfigure}
        \begin{subfigure}{\em}
                \includegraphics[width=4.5cm,height=4.2cm]{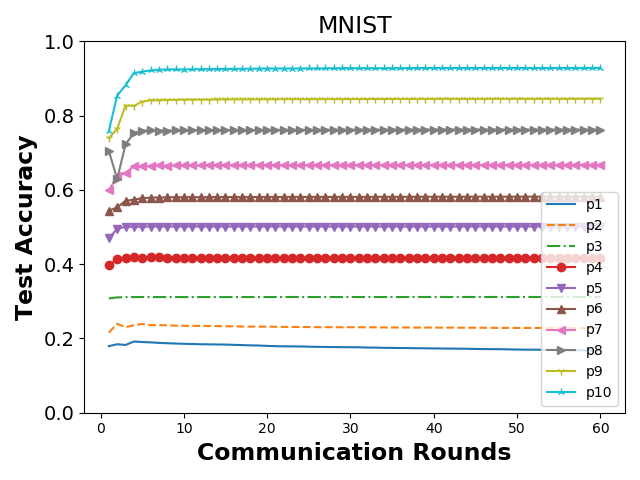}        
        \end{subfigure}
        \begin{subfigure}{\em}
                \includegraphics[width=4.5cm,height=4.2cm]{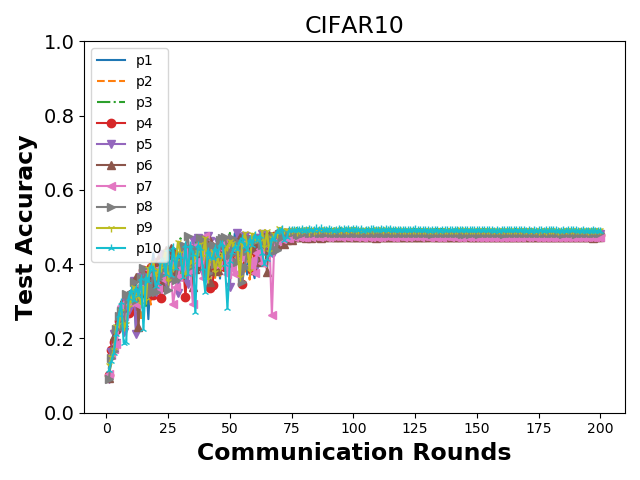}
        \end{subfigure}
        \begin{subfigure}{\em}
                \includegraphics[width=4.5cm,height=4.2cm]{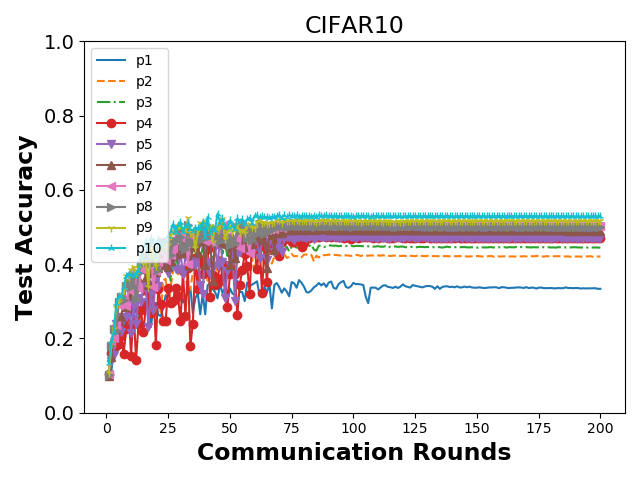}
        \end{subfigure}
        \begin{subfigure}{\em}
                \includegraphics[width=4.5cm,height=4.2cm]{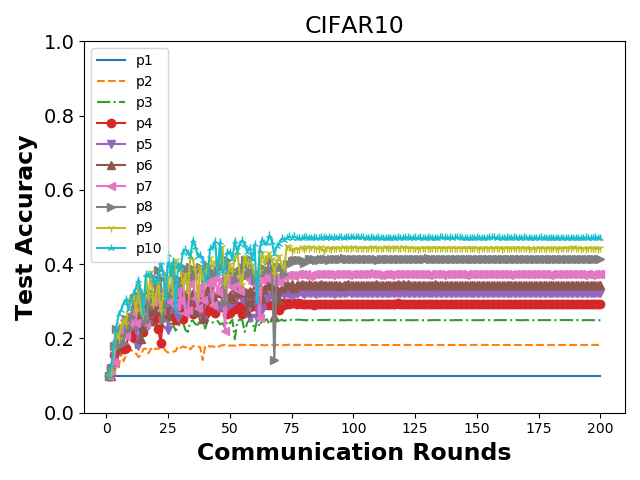}
        \end{subfigure}
         \caption{Participants final performance for MNIST and CIFAR10. From left to right \{UNI, POW, CLA\}.}
         
\label{fig:mnist_cifar10_all}
\end{figure*}

\subsection{Experimental Setup}\label{sec:A1}
\textbf{Imbalanced dataset sizes.} For CIFAR-10, we follow a power law to randomly partition total \{10000, 20000\} examples among \{5, 10\} participants respectively. For MR (SST), we follow a power law to randomly partition 9596 (8544) examples among 5 participants.

\textbf{Hyper-Parameters}. We provide the framework-independent hyperparameters used for different datasets in Table~\ref{tbl:fi_hyperparameters}. $q$-FFL: fairness coefficient $q=0.1$ and participants sampling ratio is $0.8$; SignSGD:  momentum coefficient is $0.8$ and parameter weight decay is $0.977$. FoolsGold: confidence $K=1$. Multi-Krum: participant clip ratio is $0.2$. For the hyperparameters, we either use the default values introduced in their respective papers or apply grid search to empirically find the values.

\begin{table}[ht]
\caption{Framework-independent Hyperparameters. Batch size $B$, learning rate $\eta$, exponential learning rate decay $\gamma$, total communication rounds/epochs $T$, local epochs $E$. Note that for experiments with more than 5 participants for MNIST and CIFAR-10, the learning rate $\eta$ is 0.25 and 0.025, respectively}
\label{tbl:fi_hyperparameters}
\centering
\begin{tabularx}{\linewidth}{|l|*{4}{C{1}|}}
\hline
Dataset & $B$ & $\eta$ ($\gamma$) & $T (E)$
\tabularnewline
\hline
MNIST     &   16 & 0.15~(0.977) & 60~(1)
\tabularnewline
\hline
CIFAR-10  & 64 & 0.015~(0.977) & 200~(1)
\tabularnewline
\hline
MR         &   128 & 1e-4~(0.977) & 100~(1)
\tabularnewline
\hline
SST     & 128 & 1e-4~(0.977) & 100~(1)
\tabularnewline
\hline
\end{tabularx}
\end{table}

\textbf{Runtime Statistics, Hardware and Software.}
We conduct our experiments on a machine with 12 cores~(Intel(R) Xeon(R) CPU E5-2650 v4 @ 2.20GHz), 110 GB RAM and 4 GPUs~(P100 Nvidia). Execution time for the experiments including only RFFL~(all) frameworks: for MNIST (10 participants) approximately 0.6~(0.7) hours; for CIFAR-10 (10 participants) approximately 0.7~(4.3) hours; for MR and SST (5 participants) approximately 1.5~(2) hours.

Our implementation mainly uses PyTorch, torchtext, torchvision and some auxiliary packages such as Numpy, Pandas and Matplotlib. The specific versions and package requirements are provided together with the source code. To reduce the impact of randomness in the experiments, we adopt several measures: fix the model initilizations~(we initialize model weights and save them for future experiments); fix all the random seeds; and invoke the deterministic behavior of PyTorch. As a result, given the same model initialization, 
our implementation is expected to produce consistent results on the same machine over experimental runs.

\subsection{Experimental Results} \label{sec:A2}
Comprehensive experimental results below demonstrate that RFFL is the \emph{only} framework that performs consistently well over all the investigated situations, though may not perform the best in all of them.

\textbf{5-participant Case for MNIST and CIFAR-10.} We include the fairness and accuracy results for the 5-participant case for MNIST and CIFAR-10 under the three data splits in Tables~\ref{tbl:mnist_cifar10_p5_accuracy} and~\ref{tbl:mnist_cifar10_p5_fairness}, respectively.

\begin{figure*}[!htp]
\centering
        \begin{subfigure}{\em}
                \includegraphics[width=6cm,height=4.2cm]{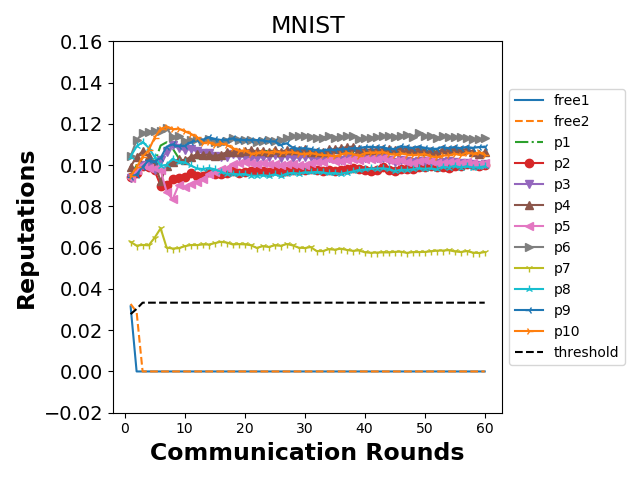}
        \end{subfigure}
        \begin{subfigure}{\em}
                \includegraphics[width=6cm,height=4.2cm]{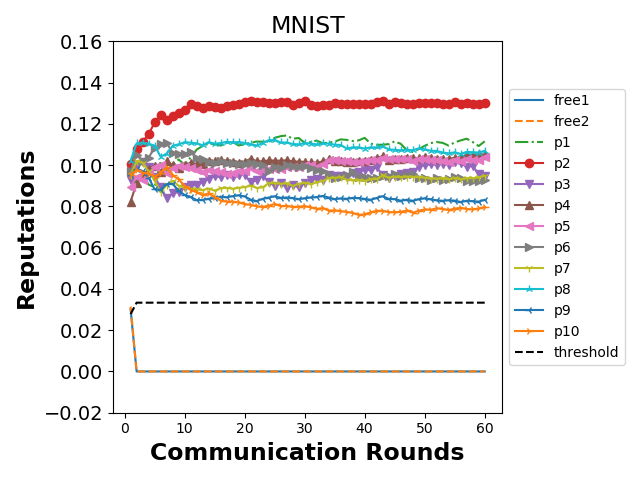}        
        \end{subfigure}
         \caption{Reputations of the participants including free-riders for the UNI~(left) and POW~(right) splits in RFFL. The reputations of these two free-riders are very quickly decayed lower than the reputation threshold, thus free-riders are identified and isolated from the system at the beginning.}
\label{fig:mnist_fr_reputations}
\end{figure*}

\textbf{Free-riders.} For better illustration and coherence, we include here the experimental results together with the participants' reputation curves. Table~\ref{tbl:untargeted_freerider_uniform} demonstrates the performance results for 20\% free-riders in the 10-participant case for MNIST over UNI. Figure~\ref{fig:mnist_fr_reputations} demonstrates the reputations of the participants. It can be clearly observed that free-riders are isolated from the federated system at the early stages of collaboration (within 5 rounds).

\begin{table}[ht]
\caption{Additional predictive performance results. Average and Maximum Test Accuracy[\%]. Values in brackets denote maximum accuracy among the participants.
}
\small
\label{tbl:mnist_cifar10_p5_accuracy}
\centering
\begin{tabularx}{\linewidth}{|l|*{6}{C{1}|}}
\hline
\multirow{2}{*}{} & \multicolumn{3}{c|}{MNIST} & \multicolumn{3}{c|}{CIFAR-10}
\tabularnewline
\cline{2-7}
 & UNI & POW & CLA & UNI & POW & CLA
\tabularnewline
\hline
\textit{Standalone} 
&  91(91) &  87(94) &  50(91)
&  44(46) &  42(52) &  29(44)
\tabularnewline
\hline
FedAvg
&  93(93) &  91(95) &  50(92)
&  46(47) &  46(52) &  30(45)
\tabularnewline
$q$-FFL
& 82(85) &  59(78) &  49(84)
& 31(32) &  31(34) &  19(24)
\tabularnewline
CFFL
& 24(39) &  21(37) &  27(28) 
& 44(45) &  40(49) &  26(43)
\tabularnewline
\hline
RFFL 
&  \textbf{97}(\textbf{97}) &  \textbf{96}(\textbf{97}) &  \textbf{79}(\textbf{94})
&  \textbf{57}(\textbf{57}) &  \textbf{56}(\textbf{58}) &  \textbf{31}(\textbf{48})
\tabularnewline
\hline
\end{tabularx}
\end{table}

\begin{table}[ht]
\caption{Additional Fairness results[\%] as in Definition~\ref{definition:collaborative-fairness} between the standalone performance and the final performance. 
}
\label{tbl:mnist_cifar10_p5_fairness}
\small
\centering
\begin{tabularx}{\linewidth}{|l|*{6}{C{1}|}}
\hline
\multirow{2}{*}{} & \multicolumn{3}{c|}{MNIST} & \multicolumn{3}{c|}{CIFAR-10}
\tabularnewline
\cline{2-7}
 & UNI & POW & CLA & UNI & POW & CLA
\tabularnewline
\hline
FedAvg
& 20.27 & 95.10 & 55.86 
& 16.92 & 84.76 & 86.20
\tabularnewline
$q$-FFL
&   66.49 &  $-$38.48 &  $-$54.85
&   17.23 &   60.47 &   28.07
\tabularnewline
CFFL
&   30.76 &   18.06 &  $-$23.04
&   66.21 &   63.35 &  $-$13.94 
\tabularnewline
RFFL
& \textbf{85.12} & \textbf{98.45} & \textbf{99.64}
& \textbf{95.99} & \textbf{99.58} & \textbf{99.93}
\tabularnewline
\hline
\end{tabularx}
\end{table}

\textbf{Adversarial Experiments with the POW.} We conduct experiments with adversaries under two data splits, the UNI and the POW. We have included the experimental results with respect to the UNI in the main paper and supplement here the experimental results with respect to the POW. Table~\ref{tbl:targeted_poisoning_powerlaw}, Table~\ref{tbl:untargeted_sign_powerlaw}, Table~\ref{tbl:untargeted_rescaling_powerlaw},
Table~\ref{tbl:untargeted_inverting_powerlaw} and Table~\ref{tbl:untargeted_freerider_powerlaw} show the respective results for the targeted poisoning adversaries, three untargeted poisoning adversaries and free-riders.

\textbf{Adversarial Experiments with Adversaries as the Majority.} For extension, we also conduct experiments by increasing the number of adversaries to test RFFL's Byzantine tolerance. Our experimental results in
 Table~\ref{tbl:targeted_poisoning_11},
 Table~\ref{tbl:untargeted_sign_uniform_11}, Table~\ref{tbl:untargeted_rescale_uniform_11}, Table~\ref{tbl:untargeted_invert_uniform_11}, and Table~\ref{tbl:untargeted_freerider_uniform_11} demonstrate that RFFL consistently achieves competitive performance over various types of adversaries even when the adversaries are the majority in the system.

\begin{figure*}[!t]
\centering
        \begin{subfigure}{\em}
                \includegraphics[width=4.5cm,height=4.2cm]{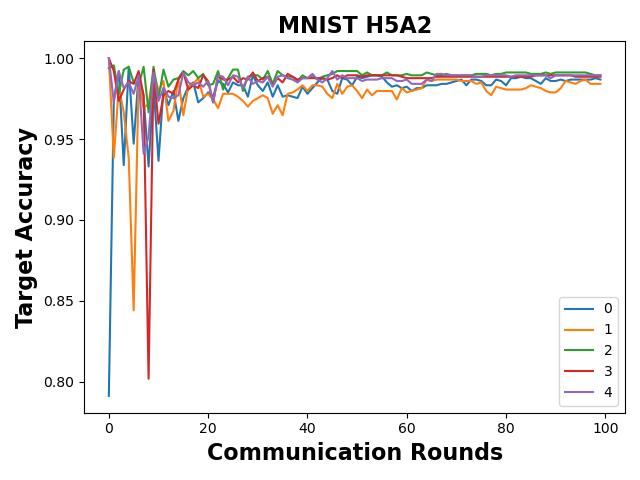}        
        \end{subfigure}
        \begin{subfigure}{\em}
                \includegraphics[width=4.5cm,height=4.2cm]{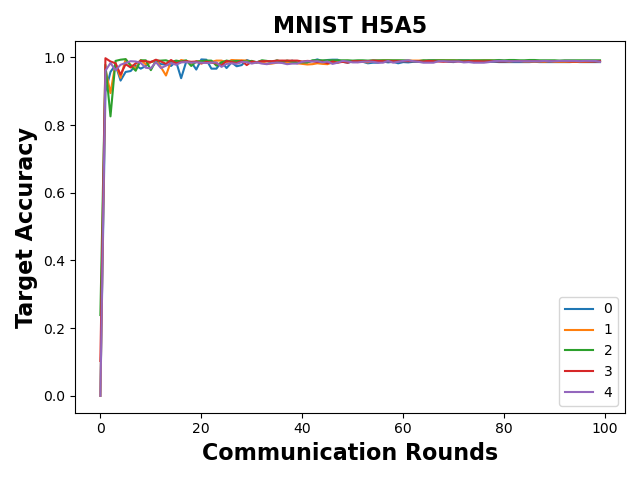}        
        \end{subfigure}
        \begin{subfigure}{\em}
                \includegraphics[width=4.5cm,height=4.2cm]{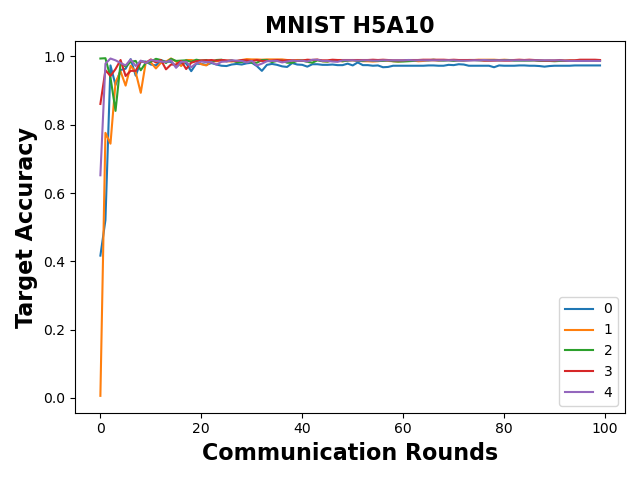}        
        \end{subfigure}

        \begin{subfigure}{\em}
                \includegraphics[width=4.5cm,height=4.2cm]{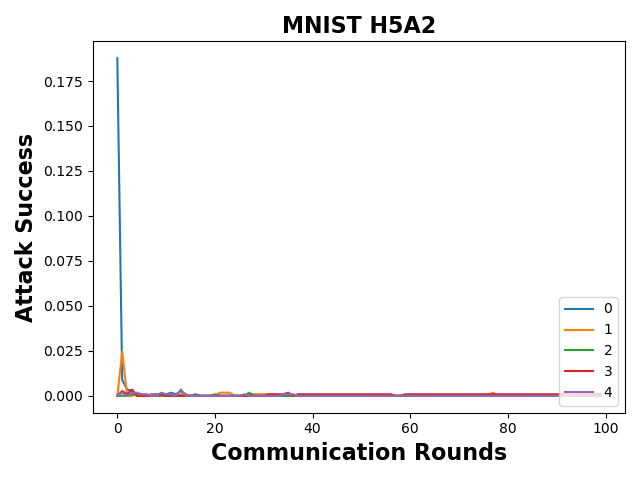}        
        \end{subfigure}
        \begin{subfigure}{\em}
                \includegraphics[width=4.5cm,height=4.2cm]{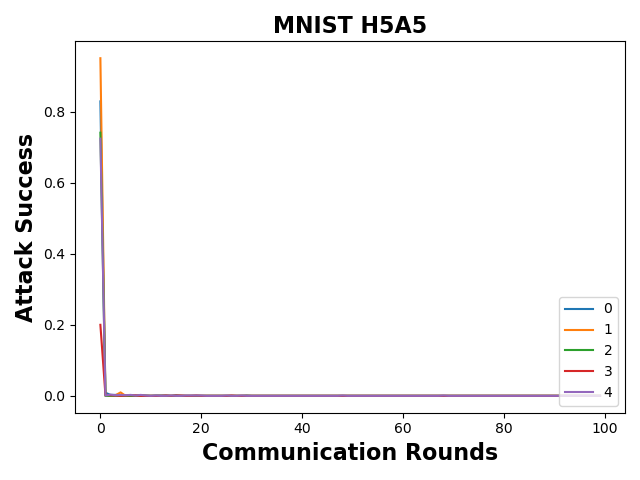}        
        \end{subfigure}
        \begin{subfigure}{\em}
                \includegraphics[width=4.5cm,height=4.2cm]{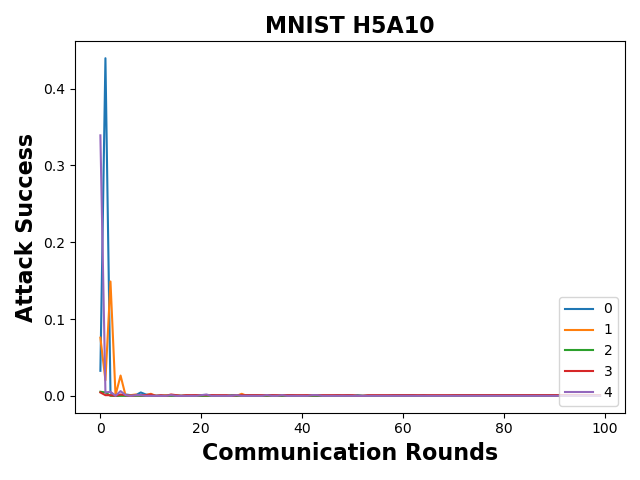}        
        \end{subfigure}
        
        \begin{subfigure}{\em}
                \includegraphics[width=4.5cm,height=4.2cm]{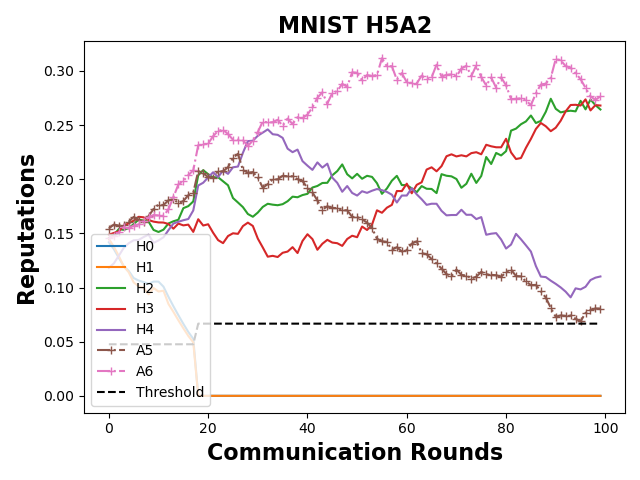}        
        \end{subfigure}
        \begin{subfigure}{\em}
                \includegraphics[width=4.5cm,height=4.2cm]{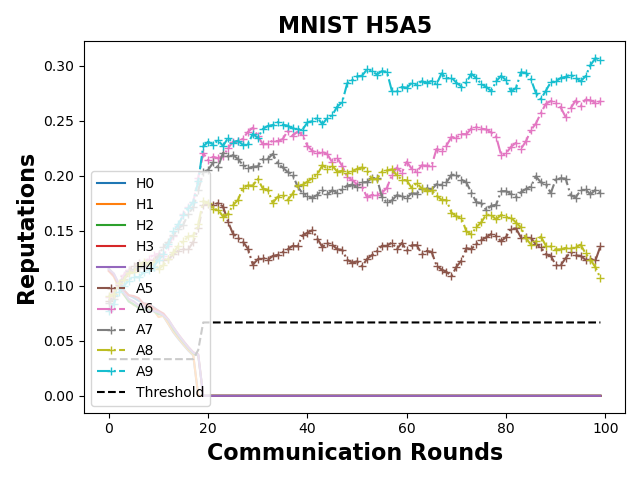}        
        \end{subfigure}
        \begin{subfigure}{\em}
                \includegraphics[width=4.5cm,height=4.2cm]{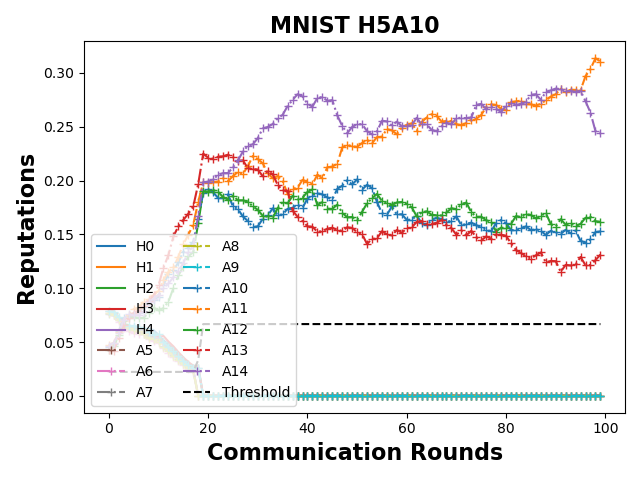}        
        \end{subfigure}
        
         \caption{Target test accuracy, attack success rate, and reputations for MNIST 5 honest participants under UNI split with \{2,5,10\} label-flipping adversaries. H5A2 refers to the FL system with 5 honest participants and 2 label-flipping adversaries.
         }
\label{fig:UNI-lf-adversaries}
\end{figure*}

\begin{figure*}[!t]
\centering
        \begin{subfigure}{\em}
                \includegraphics[width=4.5cm,height=4.2cm]{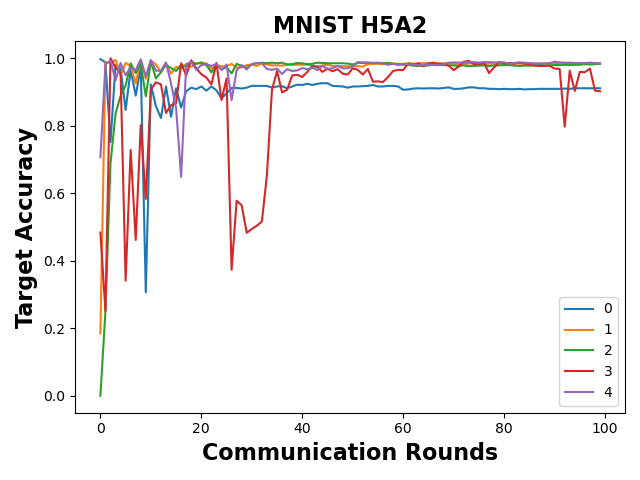}        
        \end{subfigure}
        \begin{subfigure}{\em}
                \includegraphics[width=4.5cm,height=4.2cm]{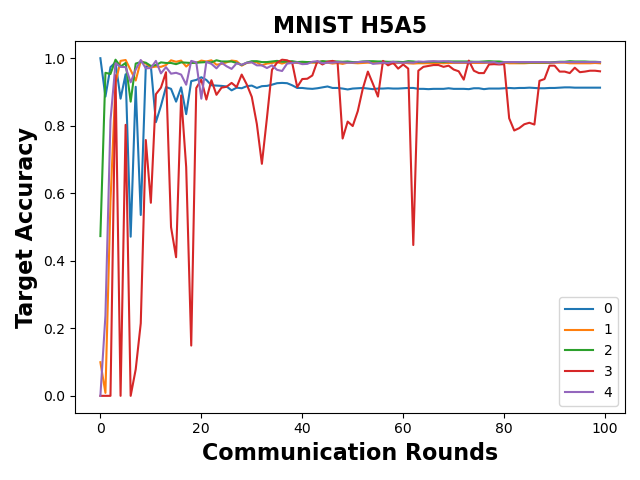}        
        \end{subfigure}
        \begin{subfigure}{\em}
                \includegraphics[width=4.5cm,height=4.2cm]{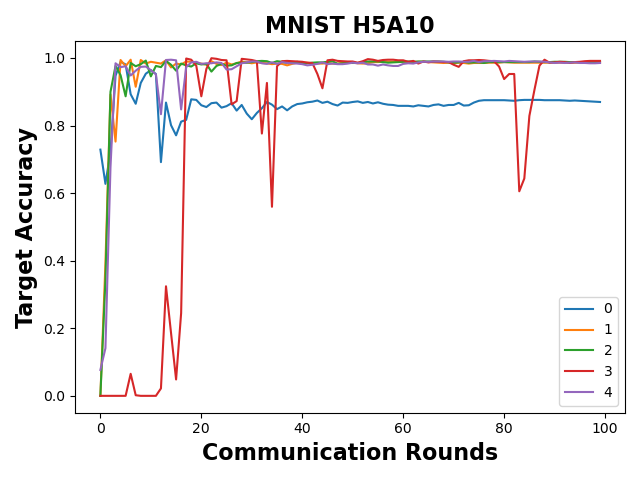}        
        \end{subfigure}

        \begin{subfigure}{\em}
                \includegraphics[width=4.5cm,height=4.2cm]{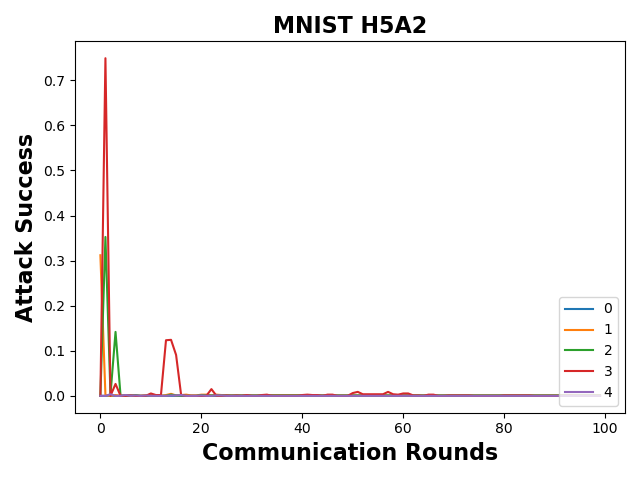}        
        \end{subfigure}
        \begin{subfigure}{\em}
                \includegraphics[width=4.5cm,height=4.2cm]{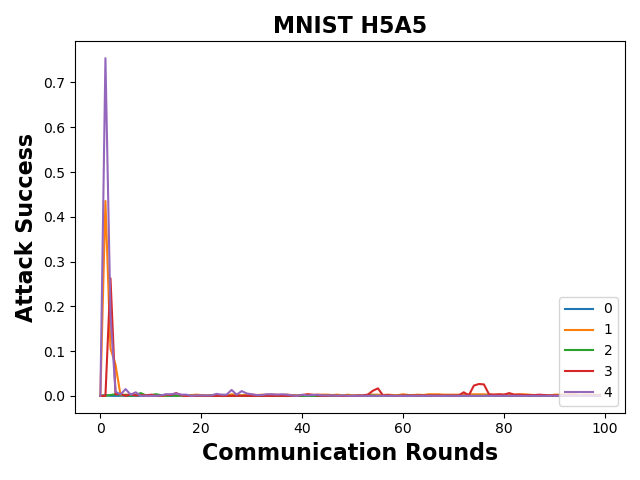}        
        \end{subfigure}
        \begin{subfigure}{\em}
                \includegraphics[width=4.5cm,height=4.2cm]{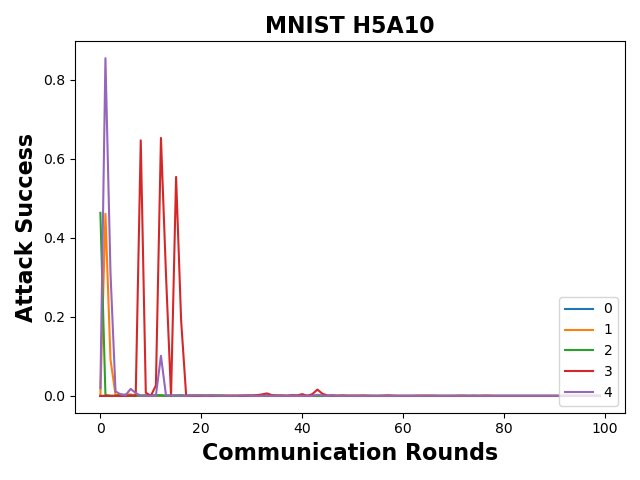}        
        \end{subfigure}
        
        \begin{subfigure}{\em}
                \includegraphics[width=4.5cm,height=4.2cm]{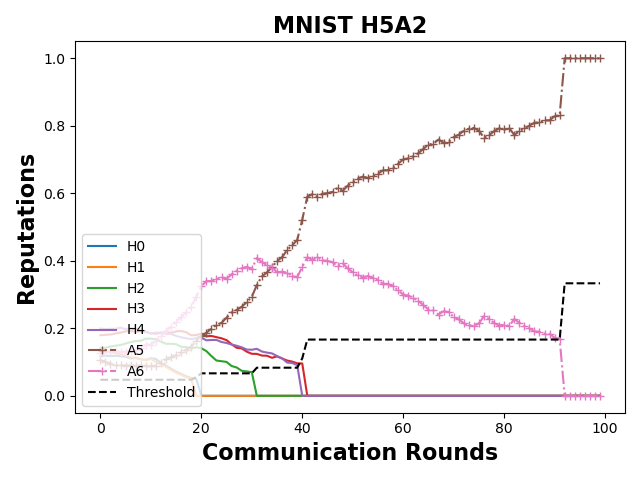}        
        \end{subfigure}
        \begin{subfigure}{\em}
                \includegraphics[width=4.5cm,height=4.2cm]{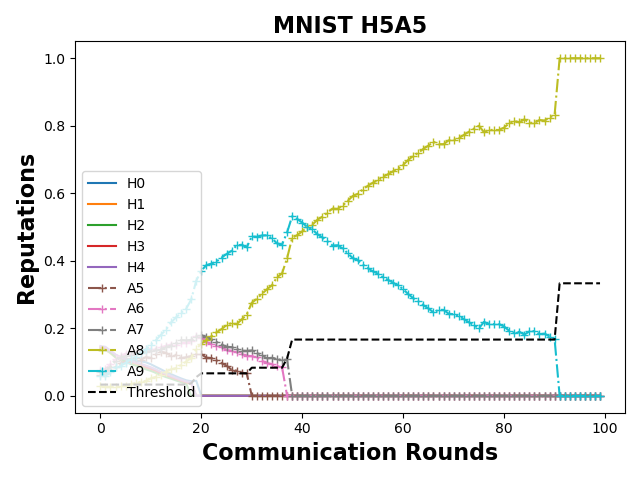}        
        \end{subfigure}
        \begin{subfigure}{\em}
                \includegraphics[width=4.5cm,height=4.2cm]{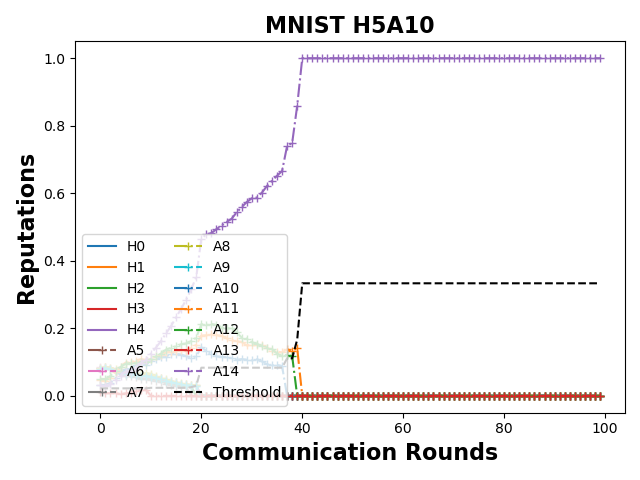}        
        \end{subfigure}
        
         \caption{Target test accuracy, attack success rate, and reputations for MNIST 5 honest participants under POW split with \{2,5,10\} label-flipping adversaries. H5A2 refers to the FL system with 5 honest participants and 2 label-flipping adversaries.
         }
\label{fig:POW-lf-adversaries}
\end{figure*}

\textbf{Additional investigation on label-flipping adversaries being the majority.}
We conduct more experiments and take a closer look at the reputations of the participants~(both honest and adversaries) over the communication rounds to gauge a better understanding of what the algorithm is doing. Specifically, we consider both the UNI and POW splits for 5 honest participants on MNIST. We vary the number of adversaries~(i.e., label-flipping attacks) from \{2,5,10\}. Each adversary has 600 I.I.D examples. We plot the target test accuracy, attack success rate and reputations over the communication rounds. Here we only plot the target test accuracy and attack success rate for honest participants. Figures~\ref{fig:UNI-lf-adversaries} and~\ref{fig:POW-lf-adversaries} show high target accuracy and low attack success (in their first and second rows), which are consistent with our tabulated results. However, in the third row of both Figures~\ref{fig:UNI-lf-adversaries} and~\ref{fig:POW-lf-adversaries}, the reputations of the honest participants~(solid lines) quickly decrease and fall below the threshold and almost all the plots end up with only adversaries~(dashed-dot lines with markers) in the system. Naturally the honest participants do not get affected by the adversaries after they are removed from the system.

These results may imply a limitation of the algorithm to defend against more sophisticated adversaries whose gradients may be similar to those honest participants with relatively poor local data. We hypothesize that further investigation of the difference between the honest and adversarial gradients would help, and leave this to future work.

\begin{table}[!htp]
\caption{Individual test accuracies [\%] over MNIST under UNI with 10 honest participants and additional 20\% \textbf{free-riders}. Free-riders omitted.}
\centering
\label{tbl:untargeted_freerider_uniform}
\begin{tabularx}{\linewidth}{|l|*{11}{C{1}|}}
\hline
{} &  1 &  2 &  3 &  4 &  5 &  6 &  7 &  8 &  9 & 10 \\
\hline
FedAvg     & 97 & 97 & 97 & 97 & 97 & 97 & 97 & 97 & 97 &  97 \\
\hline
FoolsGold  & 11 & 11 & 10 & 11 & 10 & 10 & 11 & 11 & 10 &  11 \\
Multi-Krum & 61 & 61 & 64 & 57 & 60 & 62 & 62 & 60 & 62 &  57 \\
SignSGD    &  9 &  9 &  9 &  9 &  9 &  9 &  9 &  9 &  9 &   9 \\
Median     &  1 &  1 &  1 &  1 &  1 &  1 &  1 &  1 &  1 &   1 \\
\hline
RFFL       & 92 & 93 & 94 & 92 & 93 & 93 & 91 & 92 & 93 &  92 \\
\hline
\end{tabularx}
\end{table}

\begin{table}[ht]
\caption{Maximum accuracy [\%], Attack success rate [\%] and Target accuracy [\%] over MNIST under POW with 10 honest participants and additional 20\% \textbf{label-flipping} adversaries.}
\label{tbl:targeted_poisoning_powerlaw}
\centering
\begin{tabularx}{\linewidth}{|l|*{3}{C{1}|}}
\hline
{} &  Max accuracy & Attack success rate & Target accuracy \\
\hline
FedAvg     &         \textbf{97.22} &                 0.20 &            98.80 \\
\hline
SignSGD    &          9.11 &                41.90 &            18.80 \\
FoolsGold  &          9.80 &                 \textbf{0} &             0.00 \\
Multi-Krum &         96.13 &                 \textbf{0} &            \textbf{98.90} \\
Median     &          0.09 &                 0.20 &             0.20 \\
\hline
RFFL       &         95.01 &                 \textbf{0} &            98.70 \\
\hline
\end{tabularx}
\end{table}

\begin{table}[ht]
\caption{Individual test accuracies [\%] over MNIST under POW with 10 honest participants and additional 20\% \textbf{sign-randomizing} adversaries. Adversaries omitted.}
\centering
\label{tbl:untargeted_sign_powerlaw}
\begin{tabularx}{\linewidth}{|l|*{11}{C{1}|}}
\hline
{} &  1 &  2 &  3 &  4 &  5 &  6 &  7 &  8 &  9 & 10 \\
\hline
FedAvg     & 97 & 97 & 97 & 97 & 97 & 97 & 97 & 97 & 97 &  97 \\
\hline
SignSGD    &  9 &  9 &  9 &  9 &  9 &  9 &  9 &  9 &  9 &   9 \\
FoolsGold  & 80 & 78 & 81 & 83 & 84 & 86 & 86 & 87 & 87 &  88 \\
Multi-Krum & 96 & 96 & 96 & 96 & 96 & 96 & 96 & 96 & 96 &  97 \\
Median     &  1 &  1 &  1 &  1 &  1 &  1 &  1 &  1 &  1 &   1 \\
\hline
RFFL       & 86 & 88 & 91 & 92 & 93 & 93 & 94 & 94 & 95 &  94 \\
\hline
\end{tabularx}
\end{table}

\begin{table}[ht]
\caption{Individual test accuracies [\%] over MNIST under POW with 10 honest participants and additional 20\% \textbf{re-scaling} adversaries. Adversaries omitted.}
\centering
\label{tbl:untargeted_rescaling_powerlaw}
\begin{tabularx}{\linewidth}{|l|*{11}{C{1}|}}
\hline
{} &  1 &  2 &  3 &  4 &  5 &  6 &  7 &  8 &  9 & 10 \\
\hline
FedAvg     & 10 & 10 & 10 & 10 & 10 & 10 & 10 & 10 & 10 &  10 \\
\hline
SignSGD    &  9 &  9 &  9 &  9 &  9 &  9 &  9 &  9 &  9 &   9 \\
FoolsGold  & 93 & 93 & 93 & 93 & 93 & 93 & 93 & 93 & 93 &  93 \\
Multi-Krum & 10 & 10 & 10 & 10 & 10 & 10 & 10 & 10 & 10 &  10 \\
Median     &  1 &  1 &  1 &  1 &  1 &  1 &  1 &  1 &  1 &   1 \\
\hline
RFFL       & 86 & 88 & 92 & 92 & 93 & 93 & 94 & 94 & 95 &  94 \\
\hline
\end{tabularx}
\end{table}

\begin{table}[!ht]
\caption{Individual test accuracies [\%] over MNIST under POW with 10 honest participants and additional 20\% \textbf{value-inverting} adversaries. Adversaries omitted.}
\centering
\label{tbl:untargeted_inverting_powerlaw}
\begin{tabularx}{\linewidth}{|l|*{11}{C{1}|}}
\hline
{} &  1 &  2 &  3 &  4 &  5 &  6 &  7 &  8 &  9 & 10 \\
\hline
FedAvg     & 10 & 10 & 10 & 10 & 10 & 10 & 10 & 10 & 10 &  10 \\
\hline
SignSGD    &  9 &  9 &  9 &  9 &  9 &  9 &  9 &  9 &  9 &   9 \\
FoolsGold  & 10 & 10 & 10 & 10 & 10 & 10 & 10 & 10 & 10 &  10 \\
Multi-Krum &  9 &  9 &  9 &  9 &  9 &  9 &  9 &  9 &  9 &   9 \\
Median     &  0 &  0 &  0 &  0 &  0 &  0 &  0 &  0 &  0 &   0 \\
\hline
RFFL       & 73 & 83 & 91 & 91 & 93 & 93 & 94 & 94 & 95 &  94 \\
\hline
\end{tabularx}
\end{table}

\begin{table}[!ht]
\caption{Individual test accuracies [\%] over MNIST under POW with 10 honest participants and additional 20\% \textbf{free-riders}. Adversaries omitted.}
\centering
\label{tbl:untargeted_freerider_powerlaw}
\begin{tabularx}{\linewidth}{|l|*{11}{C{1}|}}
\hline
{} &  1 &  2 &  3 &  4 &  5 &  6 &  7 &  8 &  9 & 10 \\
\hline
FedAvg     & 97 & 97 & 97 & 97 & 97 & 97 & 97 & 97 & 97 &  97 \\
\hline
SignSGD    &  9 &  9 &  9 &  9 &  9 &  9 &  9 &  9 &  9 &   9 \\
FoolsGold  & 10 & 10 & 10 &  9 & 10 & 10 & 11 & 11 & 10 &  10 \\
Multi-Krum & 53 & 57 & 58 & 58 & 55 & 53 & 56 & 59 & 58 &  61 \\
Median     &  1 &  1 &  1 &  1 &  1 &  1 &  1 &  1 &  1 &   1 \\
\hline
RFFL       & 86 & 89 & 90 & 92 & 93 & 93 & 94 & 94 & 95 &  95 \\
\hline
\end{tabularx}
\end{table}

\begin{figure*}[!t]
\centering
        \begin{subfigure}{\em}
                \includegraphics[width=4.5cm,height=4.2cm]{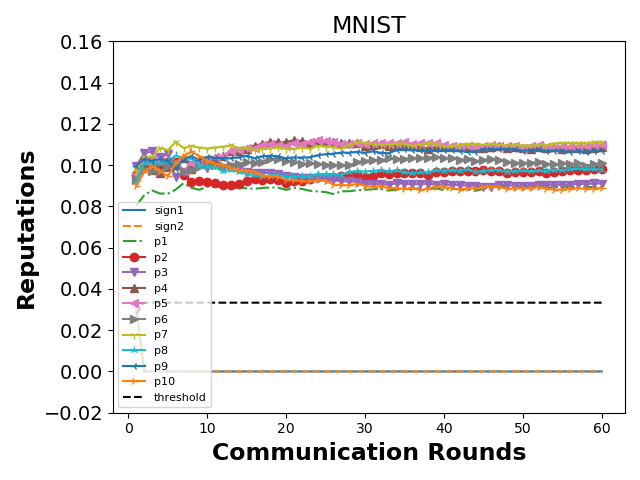}        
        \end{subfigure}
        \begin{subfigure}{\em}
                \includegraphics[width=4.5cm,height=4.2cm]{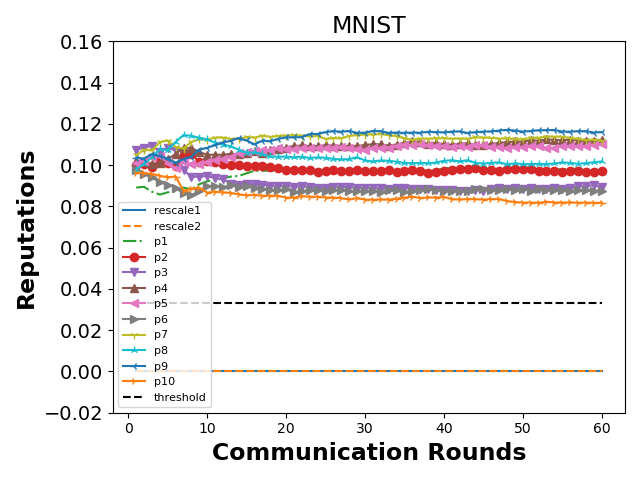}
        \end{subfigure}
        \begin{subfigure}{\em}
                \includegraphics[width=4.5cm,height=4.2cm]{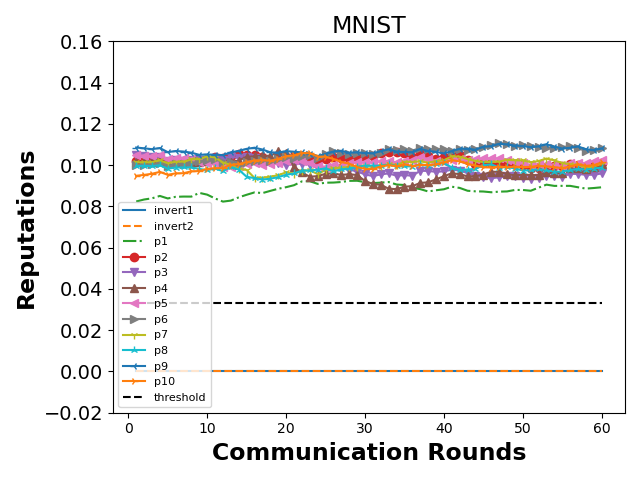}
        \end{subfigure}
         \caption{Reputations for MNIST 10 participants with 2 adversaries of untargeted poisoning. From left to right, \{\textbf{sign-randomizing}, \textbf{re-scaling}, \textbf{value-inverting}\}. The adversaries have clearly lower reputations and are removed.}
\label{fig:mnist_ma_reputations}
\end{figure*}

\begin{table}[!htp]
\caption{Individual test accuracies [\%] over MNIST under UNI with 10 honest participants and additional 110\% \textbf{sign-randomizing} adversaries. 10 honest participants with 11 adversaries.}
\centering
\label{tbl:untargeted_sign_uniform_11}
\begin{tabularx}{\linewidth}{|l|*{11}{C{1}|}}
\hline
{} &  1 &  2 &  3 &  4 &  5 &  6 &  7 &  8 &  9 & 10 \\
\hline
FedAvg     & 96 & 96 & 96 & 96 & 96 & 96 & 96 & 96 & 96 &  96 \\
\hline
SignSGD    &  9 &  9 &  9 &  9 &  9 &  9 &  9 &  9 &  9 &   9 \\
FoolsGold  & 61 & 58 & 64 & 60 & 62 & 66 & 54 & 58 & 60 &  58 \\
Multi-Krum & 95 & 94 & 96 & 95 & 96 & 95 & 96 & 95 & 95 &  95 \\
Median     &  1 &  1 &  1 &  1 &  1 &  1 &  1 &  1 &  1 &   1 \\
\hline
RFFL       & 93 & 93 & 94 & 91 & 92 & 93 & 93 & 92 & 92 &  92 \\
\hline
\end{tabularx}
\end{table}

\begin{table}[!htp]
\caption{Individual test accuracies [\%] over MNIST under UNI with 10 honest participants and additional 110\% \textbf{re-scaling} adversaries. 10 honest participants with 11 adversaries.}
\centering
\label{tbl:untargeted_rescale_uniform_11}
\begin{tabularx}{\linewidth}{|l|*{11}{C{1}|}}
\hline
{} &  1 &  2 &  3 &  4 &  5 &  6 &  7 &  8 &  9 & 10 \\
\hline
FedAvg     & 10 & 10 & 10 & 10 & 10 & 10 & 10 & 10 & 10 &  10 \\
\hline
SignSGD    &  9 &  9 &  9 &  9 &  9 &  9 &  9 &  9 &  9 &   9 \\
FoolsGold  & 11 & 11 & 11 & 11 & 11 & 11 & 11 & 11 & 11 &  11 \\
Multi-Krum & 10 & 10 & 10 & 10 & 10 & 10 & 10 & 10 & 10 &  10 \\
Median     & 93 & 93 & 93 & 93 & 93 & 93 & 93 & 93 & 93 &  93 \\
\hline
RFFL       & 93 & 92 & 94 & 92 & 93 & 93 & 93 & 92 & 93 &  93 \\
\hline
\end{tabularx}
\end{table}

\begin{table}[!htp]
\caption{Individual test accuracies [\%] over MNIST under UNI with 10 honest participants and additional 110\% \textbf{value-inverting} adversaries. 10 honest participants with 11 adversaries.}
\centering
\label{tbl:untargeted_invert_uniform_11}
\begin{tabularx}{\linewidth}{|l|*{11}{C{1}|}}
\hline
{} &  1 &  2 &  3 &  4 &  5 &  6 &  7 &  8 &  9 & 10 \\
\hline
FedAvg     &  9 &  9 &  9 &  9 &  9 &  9 &  9 &  9 &  9 &   9 \\
\hline
SignSGD    &  9 &  9 &  9 &  9 &  9 &  9 &  9 &  9 &  9 &   9 \\
FoolsGold  & 10 & 10 & 10 & 10 & 10 & 10 & 10 & 10 & 10 &  10 \\
Multi-Krum & 18 & 18 & 18 & 18 & 18 & 18 & 18 & 18 & 18 &  18 \\
Median     &  9 &  9 &  9 &  9 &  9 &  9 &  9 &  9 &  9 &   9 \\
\hline
RFFL       & 93 & 92 & 94 & 92 & 93 & 93 & 93 & 92 & 93 &  93 \\
\hline
\end{tabularx}
\end{table}

\begin{table}[!htp]
\caption{Individual test accuracies [\%] over MNIST under UNI with 10 honest participants and additional 110\% \textbf{free-riders}. 10 honest participants and 11 free-riders.}
\centering
\label{tbl:untargeted_freerider_uniform_11}
\begin{tabularx}{\linewidth}{|l|*{11}{C{1}|}}
\hline
{} &  1 &  2 &  3 &  4 &  5 &  6 &  7 &  8 &  9 & 10 \\
\hline
FedAvg     & 97 & 97 & 97 & 97 & 97 & 97 & 97 & 97 & 97 &  97 \\
\hline
SignSGD    &  9 &  9 &  9 &  9 &  9 &  9 &  9 &  9 &  9 &   9 \\
FoolsGold  & 10 & 10 & 10 & 10 & 10 & 10 & 10 & 10 & 10 &  10 \\
Multi-Krum & 51 & 52 & 45 & 46 & 41 & 47 & 43 & 46 & 47 &  47 \\
Median     &  1 &  1 &  1 &  1 &  1 &  1 &  1 &  1 &  1 &   1 \\
\hline
RFFL       & 92 & 94 & 93 & 92 & 92 & 93 & 93 & 93 & 92 &  92 \\
\hline
\end{tabularx}
\end{table}